\documentclass[runningheads]{llncs}
\usepackage[T1]{fontenc}
\usepackage{graphicx,verbatim}
\usepackage{amssymb}
\usepackage{graphicx}
\usepackage{amsmath}
\usepackage{booktabs}
\usepackage{algorithmic}
\usepackage[linesnumbered,ruled,vlined]{algorithm2e}
\usepackage[switch]{lineno}
\usepackage{multirow}
\usepackage{subcaption}
\usepackage{array}
\usepackage{color}
\usepackage{xcolor}
\usepackage{bm}
\usepackage{hyperref}
\usepackage{orcidlink}
\usepackage{fontawesome5}
\begin{document}
\title{Scalp Diagnostic System With Label-Free Segmentation and Training-Free Image Translation}
%
% \begin{comment}  %% Removed for anonymized MICCAI 2025 submission
% \author{Youngmin Kim$^{\dagger}$\inst{1}\orcidlink{0000-0002-5948-456X} \and
% Saejin Kim$^{\dagger}$\inst{1}\orcidlink{0009-0008-6502-7008} \and
% Hoyeon Moon\inst{1}\orcidlink{0009-0009-3552-1598} \and
% Youngjae Yu$^{\ddagger}$\inst{2}\orcidlink{0000-0002-5867-0782} \and
% Junhyug Noh$^{\ddagger}$\inst{3}\orcidlink{0000-0003-1239-8178}
% }

\author{Youngmin Kim$^{\dagger}$\inst{1}\orcidlink{0000-0002-5948-456X}
% index{Kim, Youngmin}
\and
Saejin Kim$^{\dagger}$\inst{1}\orcidlink{0009-0008-6502-7008}
% index{Kim, Saejin}
\and
Hoyeon Moon\inst{1}\orcidlink{0009-0009-3552-1598}
% index{Moon, Hoyeon}
\and
Youngjae Yu$^{\ddagger}$\inst{2}\orcidlink{0000-0002-5867-0782}
% index{Yu, Youngjae}
\and
Junhyug Noh$^{\ddagger}$\inst{3}\orcidlink{0000-0003-1239-8178}
% index{Noh, Junhyug}
}

\authorrunning{Kim et al.}
\institute{
Yonsei University, 50, Yonsei-ro, Seodaemun-gu, Seoul, Korea \and
Seoul National University, 1, Gwanak-ro, Gwanak-gu, Seoul, Korea \and
Ewha Womans University, 52, Ewhayeodae-gil, Seodaemun-gu, Seoul, Korea
\email{\{winston1214, jerry0110, mhy9910\}@yonsei.ac.kr}, \quad
\email{youngjaeyu@snu.ac.kr}, \quad
\email{junhyug@ewha.ac.kr}
}

% \author{Anonymized Authors}  %% Added for anonymized MICCAI 2025 submission
% \authorrunning{Anonymized Author et al.}
% \institute{Anonymized Affiliations \\
%     \email{email@anonymized.com}}
\titlerunning{ScalpVision}
\def\thefootnote{$\dagger$}\footnotetext{Equal contribution\quad$^{\ddagger}$ Co-supervision}
\maketitle              % typeset the header of the contribution

\begin{abstract}
Scalp disorders are highly prevalent worldwide, yet remain underdiagnosed due to limited access to expert evaluation and the high cost of annotation. Although AI-based approaches hold great promise, their practical deployment is hindered by challenges such as severe data imbalance and the absence of pixel-level segmentation labels. To address these issues, we propose ``\textsc{ScalpVision}'', an AI-driven system for the holistic diagnosis of scalp diseases. In \textsc{ScalpVision}, effective hair segmentation is achieved using pseudo image-label pairs and an innovative prompting method in the absence of traditional hair masking labels. Additionally, \textsc{ScalpVision} introduces \emph{DiffuseIT-M}, a generative model adopted for dataset augmentation while maintaining hair information, facilitating improved predictions of scalp disease severity. Our experimental results affirm \textsc{ScalpVision}'s efficiency in diagnosing a variety of scalp conditions, showcasing its potential as a valuable tool in dermatological care. 
Our code is available at~\faGithub\hspace{0.1em} \href{https://github.com/winston1214/ScalpVision}{winston1214/ScalpVision}.


\end{abstract}

\section{Introduction}
\label{sec:introduction}
Scalp disorders are a widespread concern, with nearly $90\%$ of adults in the U.S. experiencing some form of condition~\cite{elewski2005clinical}. 
Left unchecked, even seemingly mild scalp ailments can escalate into more serious outcomes, such as alopecia, underscoring the importance of timely intervention.
Consequently, early diagnosis is crucial for preventing the progression of scalp-related diseases~\cite{panjwani2009:early,pratt2017:alopecia}, highlighting the need for advanced diagnostic approaches that are both efficient and accessible.
Recognizing the importance of early detection, numerous studies have explored scalp disease diagnosis using microscopic scalp imagery~\cite{chang2020:scalpeye,kim2022:hair,seo2020:trichoscopy}.

Nevertheless, effectively diagnosing scalp disorders relies heavily on measuring critical features such as hair count and thickness, which demand precise hair segmentation. 
However, generating pixel-level hair annotations is costly and time-consuming, and no publicly available dataset provides such segmentation labels. 
The only major resource, AI-Hub~\cite{aihub:2020}, offers classification labels for scalp conditions but lacks segmentation annotations (see Section~\ref{sec:dataset}). 
Moreover, like many scalp image datasets, it suffers from data imbalance, especially for severe conditions, making it challenging to develop robust models.

To overcome these limitations, we propose \textsc{ScalpVision}, a comprehensive system for the in-depth assessment of scalp health. 
First, we achieve label-free hair segmentation by combining a naive segmentation model -- trained on synthetic image-label pairs -- with an \emph{automatic prompting} module for the Segment Anything Model (SAM)~\cite{kirillov2023:segment}, systematically generating positive and negative point prompts to enable accurate hair masks without manual labeling.
Building on these masks, we then introduce \emph{DiffuseIT-M}, a diffusion-based image-to-image translation framework that preserves hair details while altering scalp conditions.
By generating diverse training samples, our method effectively mitigates data imbalance, ultimately leading to enhanced diagnostic performance for scalp diseases.

\begin{figure*}[t!]
    \centering
        \centering
        \includegraphics[width=\linewidth]{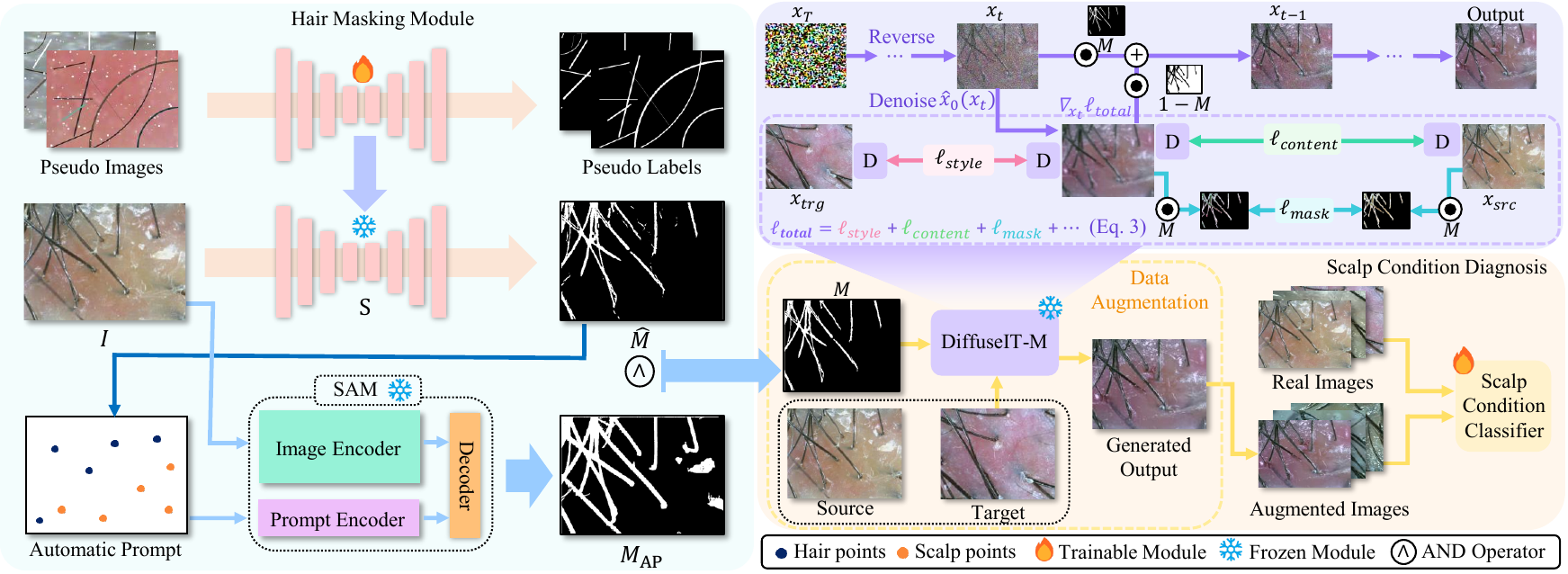}
        \caption{ 
        \textsc{ScalpVision} pipeline overview: $I$ is the original image, model $S$ generates the hair segmentation mask $\hat{M}$ using a pseudo-training set, $M_{\text{AP}}$ is the SAM-produced mask, and $M$ is the combined hair segmentation mask. The ``Automatic Prompt'' for refining segmentation comes from $\hat{M}$. $x_{src}$ and $x_{trg}$ are the source and target images, with $M$ as the mask image of $x_{src}$. The weighted image sum is denoted by $\odot$ and $D$ stands for DINO-ViT~\cite{caron2021:emerging}.
        }
        \label{fig:detail}
        \label{fig:diffIT-M}
\end{figure*}

\section{Method}
\label{sec:method}
As illustrated in Figure~\ref{fig:detail}, central to \textsc{ScalpVision} is a hair segmentation module (Section~\ref{sec:unsuper-seg}) and an image translation module for generating diverse scalp images to augment training datasets for scalp condition classification (Section~\ref{sec:scalp_condition_classification}).

\subsection{Label-Free Hair Segmentation}
\label{sec:unsuper-seg}
For the precise diagnosis of scalp conditions, our initial step involves segmenting hair within microscopic scalp images.
However, since most scalp condition datasets lack segmentation labels, supervised learning methods are not feasible.

\noindent \textbf{Heuristic-driven pseudo-labeling.}
To address the absence of hair segmentation, we first generate pseudo labels for training our segmentation model ($\textrm{S}$ as shown in Figure~\ref{fig:detail}) using prior knowledge. 
With the intuition that the hair on the microscopic scalp images follows either a linear function or a power function, we generate synthetic images to effectively guide the model to learn hair patterns on the scalp images. 
For each disease condition, we randomly select one image representing each distinct severity level, extract three smaller patches from regions of the scalp with no visible hair, and draw curves to simulate hair patterns.
Additionally, to simulate dandruff noise, circular white shapes are added to these patches but are not indicated in the pseudo masks, thus training the model to interpret them as noise. 
Examples of the pseudo images and masks are provided in the Appendix~\ref{sec:suppl_seg}.
We generate $3{,}000$ pseudo-images and corresponding pseudo mask labels, using them to train the U$^{2}$-Net~\cite{qin2020:u2} which generates the binary mask, $\hat{M} = \hat{M}(i,j) \in \{0, 1\}^{H \times W}$,
where \(H\) and \(W\) are the height and width of the image, and \(i \in [1, H]\), \(j \in [1, W]\) denote pixel coordinates.

\begin{algorithm}[t]
\caption{Extraction of representative points from mask}
\label{algorithm:bbox_extraction}
\small
\DontPrintSemicolon
\KwIn{Mask $\hat{M}$, bounding box size $n$, cross-shaped structuring element \textit{kernel}}
\KwOut{Representative hair points from mask $\hat{\mathcal{C}}$}

$\mathcal{H}_{\text{copy}} \gets \hat{M}\,$;
$\hat{\mathcal{H}}_{\text{skel}} \gets$ zero array with same size as $\mathcal{H}_{\text{copy}}$;
$\hat{B}, \hat{\mathcal{C}} \gets \{ \}$\;

\While{$\mathcal{H}_{\text{copy}} \neq 0$}{
    $\textit{Eroded}$ $ \gets \texttt{MORPHOLOGY}\_ \texttt{ERODE}(\mathcal{H}_{\text{copy}}, \textit{kernel})$\;
    $\textit{Dilated} \gets \texttt{MORPHOLOGY}\_ \texttt{DILATE}(\textit{Eroded}, \textit{kernel})$\;
    $\hat{K} \gets \mathcal{H}_{\text{copy}} - \textit{Dilated}\,$;
    $\hat{\mathcal{H}}_{\text{skel}} \gets \hat{\mathcal{H}}_{\text{skel}} \lor \hat{K}\,$;
    $\mathcal{H}_{\text{copy}} \gets \textit{Eroded}$\;
}

\ForEach{$(x,y) \in \mathcal{H}_{\text{skel}}$}{
    $\hat{B} \gets \hat{B} \cup \{(x-\frac{1}{2}n, y-\frac{1}{2}n, x+\frac{1}{2}n, y+\frac{1}{2}n)\}$\;
}

$\hat{B} \gets \texttt{NMS}(\hat{B})\,$

\ForEach{$(x_1, y_1, x_2, y_2) \in \hat{B}$}{
    $\hat{\mathcal{C}} \gets \hat{\mathcal{C}} \cup \{ (\bar{x}, \bar{y}) \} \text{ as in Eq.(\ref{eq:mean_point})}$\;
}

\Return $\hat{\mathcal{C}}$\;
\normalsize
\end{algorithm}

\noindent \textbf{Automatic prompting for SAM.}
To refine the hair segmentation mask $\hat{M}$, we utilize the foundation segmentation model, SAM~\cite{kirillov2023:segment}, employing a point-prompting method to differentiate hair from scalp without additional training.
However, selecting random points from $\hat{M}$ for positive prompts often led to suboptimal masks, mainly due to points near the edges of $\hat{M}$ confusing the SAM.
Furthermore, the intrinsic randomness occasionally caused sampled points to coalesce within a confined region, thereby leading the SAM to segment only a limited subset of hairs.
To address these issues, we developed an automatic prompting method, shown in Algorithm~\ref{algorithm:bbox_extraction}, that uniformly samples across $\hat{M}$ and uses the coarse segmentation mask $\hat{M}$ to guide the SAM with high confidence.

To extract the distinct features of the hair, we compute the skeletonized mask, $\hat{\mathcal{H}}_{\text{skel}} \in \{0, 1\}^{H \times W}$, using morphological erosion and dilation following~\cite{Zhang1984:morphSkel}.
Then, we generate bounding boxes around each pixel in $\hat{\mathcal{H}}_{\text{skel}}$ with size $n \times n$ where we set $n=10$.
These boxes undergo non-maximum suppression (NMS) to filter out the bounding boxes, denoted as $\hat{B} = \{\hat{b}_j\}_{j=1}^{k}$, where each box is defined by coordinates $(x_{\text{min}}, y_{\text{min}}, x_{\text{max}}, y_{\text{max}})$. 
Following this, the mean points of the hair pixels, $\hat{\mathcal{C}}=\{\hat{c}_j\}_{j=1}^{k}$, in each bounding box $\hat{B}$ can be determined. 
For each $\hat{b}_j = (x_1, y_1, x_2, y_2)$, the mean point $\hat{c}_j = (\bar{x}, \bar{y})$ is given by:
\begin{equation}
    \bar{x} = \frac{\sum_{i} \sum_{j} i \cdot \hat{\mathcal{H}}(i,j)}{\sum_{i} \sum_{j} \hat{\mathcal{H}}(i,j)},
    \quad
    \bar{y} = \frac{\sum_{i} \sum_{j} j \cdot \hat{\mathcal{H}}(i,j)}{\sum_{i} \sum_{j} \hat{\mathcal{H}}(i,j)}
    \label{eq:mean_point}
\end{equation}
where the summation is over all $i \in [x_1, x_2]$ and $j \in [ y_1, y_2]$. 

Subsequently, we select positive point prompts for SAM from the calculated mean points $\hat{\mathcal{C}}$. 
For the negative point prompts, we utilize the inverse of the initial mask, specifically $1 - \hat{M}$.
These prompts, automatically generated, guide SAM in generating the binary segmentation mask, $M_{\text{AP}} \in \{0, 1\}^{H \times W}$.

\noindent \textbf{Mask ensemble.}
$M_\text{AP}$ and $\hat{M}$ complement each other with strengths and weaknesses. $\hat{M}$ is robust against noise like dandruff as it was trained using simulated noise. Meanwhile, $M_\text{AP}$, benefiting from SAM's superior edge detection, excels in constructing a clear boundary between hair and scalp.
Therefore, to make a robust hair mask, the final binary mask, $M$, is derived from $\hat{M}$ and $M_{\text{AP}}$ with the logical AND operation ($M = \hat{M} \land M_{\text{AP}}$), followed by a noisy region removal post-processing step with connected-component analysis.

\subsection{Scalp Condition Classification}
\label{sec:scalp_condition_classification}
Accurately classifying scalp disease severity from microscopic images is difficult due to the rarity of extreme cases. 
To address this, we introduce \emph{DiffuseIT-M}, a diffusion-based image translation model with mask guidance that transforms a source image into various scalp conditions while preserving hair content. 
Building on DiffuseIT~\cite{kwon2022:diffusion} and incorporating an image editing technique inspired by blended diffusion~\cite{avrahami2022:blended}, \emph{DiffuseIT-M} enables robust augmentation of underrepresented classes for improved classification.

\noindent\textbf{Image translation with mask guidance.}
To facilitate the transfer of scalp disease characteristics while preserving hair features in our model, we utilize a comprehensive loss function, $\ell_{total}$, that guides the reverse process and is composed of five distinct loss components. 
These components consider the source image ($x_{src}$), the target image ($x_{trg}$), and the hair mask ($M$) as inputs. 
The combined loss function is defined as:
\begin{equation}
    \begin{gathered}
    \ell_{total} \left( \scriptstyle{x; x_{src}, x_{trg}, M} \right) =
    \lambda_{1}\ell_{style} + \lambda_{2}\ell_{content} + \lambda_{3}\ell_{mask} + \lambda_{4}\ell_{sem} + \lambda_{5}\ell_{rng},
    \end{gathered}
    \label{eq:diffit_loss}
\end{equation}
where $\lambda_{i\in [1,5]}$ denotes the weights assigned to each of these loss functions.

For $\ell_{style}$ and $\ell_{content}$, we utilize the style and content loss functions from DiffuseIT, with further details provided in the Appendix~\ref{sec:appendix_details}.
We employ the \texttt{[CLS]} token matching loss using DINO-ViT~\cite{caron2021:emerging} to reflect semantic information in $x_{trg}$ and use keys of multi-head self-attention layers to preserve the content of $x_{src}$.
Additionally, to ensure hair preservation while translating scalp styles, we construct a mask preservation loss function as:
\begin{equation}
    \ell_{mask} = \mbox{\small \textit{LPIPS}}(x_{src} \odot M, \hat{x}_{0}(x_{t}) \odot M) 
    + || (x_{src} - \hat{x}_{0}(x_{t})) \odot M ||_2,
\end{equation}
where \textit{LPIPS} denotes the learned perceptual image patch similarity metric~\cite{zhang2018:unreasonable} and $\hat{x}_{0} (x_t)$ is the estimation of the cleaned image derived from the sample $x_t$:
\begin{equation} 
    \hat{x}_{0}(x_t) = \frac{x_{t}}{\sqrt{\bar{\alpha}_{t}}} - \frac{\sqrt{1 - \bar{\alpha}_{t}}\epsilon_{\theta}(x_t , t)}{\sqrt{\bar{\alpha}_{t}}}. 
\end{equation}

We also include two additional losses: 
$\ell_{rng}$, representing the squared spherical distance as proposed in~\cite{crowson2022:vqgan}, 
and $\ell_{sem}$, indicating the semantic divergence loss as outlined in~\cite{kwon2022:diffusion}.
Using this composite loss function, $\ell_{total}$, we guide the generation of the next sample step, $x_{t-1}$. 
To preserve hair details, we apply a masking approach: 
\begin{equation} 
    x_{t-1} \leftarrow x_{t} \odot M + \bigl[\hat{x}_{0} (x_t) - \nabla{x_t}\ell_{total}\bigl(\hat{x}_{0}(x_t)\bigr)\bigr] \odot (1-M).
\end{equation}
This method allows scalp style translation without extra training.

\noindent\textbf{Classification strategy.}
Using \emph{DiffuseIT-M}, we augment our training set by translating randomly chosen images into higher severity levels via weighted sampling, where selection probability is inversely proportional to the class's size. The augmentation strategy and the corresponding results are provided in the Appendix~\ref{sec:suppl_scalp_class}. We fine-tune a pretrained backbone with four MLP heads, each tied to a specific loss. One head detects the presence of scalp diseases (dandruff, excess sebum, erythema), while the other three classify their severities (good, mild, moderate, severe). Our objective, $\ell_{cls}$, is the sum of four losses: $\ell_{dis}$ (binary cross-entropy for disease presence), and $\ell_{dand}$, $\ell_{seb}$, $\ell_{ery}$(cross-entropy for severity classification). This design enables simultaneous disease detection and severity assessment.

\begin{figure*}[t!]
    \centering
    \includegraphics[width=\linewidth]{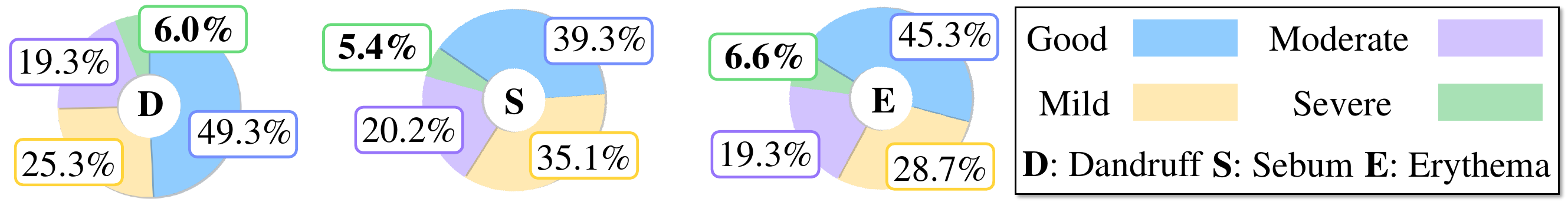}
    \caption{Data distribution of different severity within each scalp condition. 
    }
    \label{fig:data_stat}
\end{figure*}

\section{Experiments}
\label{sec:experiments}

\subsection{Dataset}
\label{sec:dataset}

Despite the lack of publicly available datasets, we accessed a specialized dataset from AI-Hub~\cite{aihub:2020} for classifying the severity of scalp dermatologic conditions. 
\footnote{This dataset is provided by `The Open AI Dataset Project (AI-Hub, S. Korea)' and is exempt from IRB approval as it does not contain any information that can identify individuals. The dataset is publicly accessible at \url{https://aihub.or.kr}.}
The dataset comprises $95{,}910$ images with a resolution of $640 \times 480$ pixels from $20{,}000$ patients. 
It is split into $72{,}342$ training and $23{,}568$ test images, with $21{,}703$ from the training set used for validation.
Dermatologists labeled each image for \textit{dandruff}, \textit{excess sebum}, and \textit{erythema}, categorizing severity as \textit{good}, \textit{mild}, \textit{moderate}, or \textit{severe}.
The dataset is heavily skewed toward \textit{good} and \textit{mild} cases (Fig.~\ref{fig:data_stat}) and lacks segmentation labels.
Thus, we manually annotated hair regions in $150$ test images to evaluate our hair segmentation methods.

\begin{figure*}[t!]
    \centering
    \includegraphics[width=\linewidth]{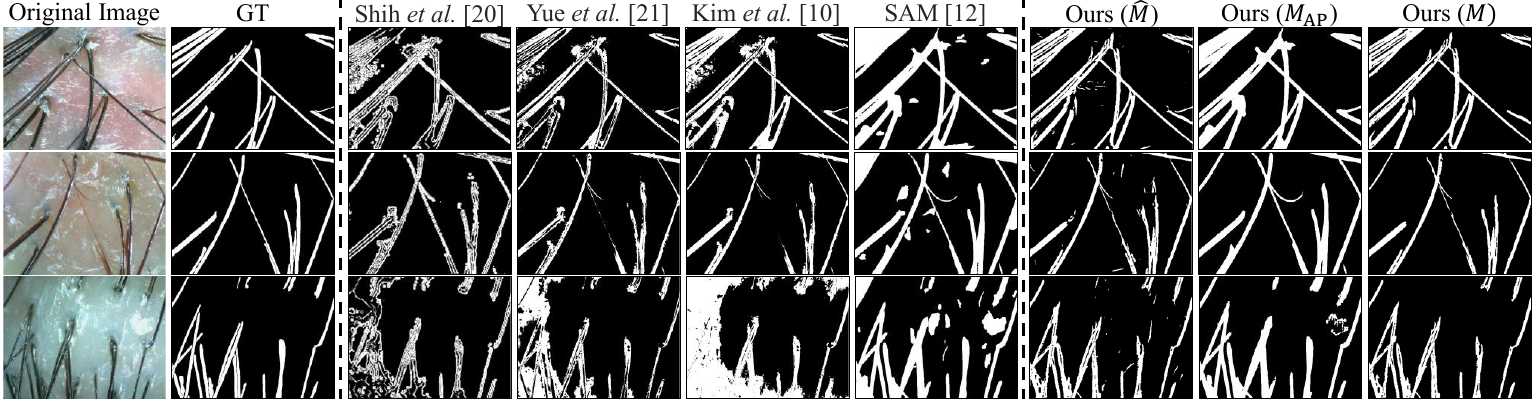}
    \caption{Comparison of various segmentation methods on hair. ``GT'' represents the mask images for which we have manually annotated the pixel segmentation. Note that $\hat{M}$, $M_\text{AP}$ and $M$ are proposed in Section~\ref{sec:unsuper-seg}. }
    \label{fig:comp_seg}
\end{figure*}

%this is 3s.f version
\begin{table}[!t]
\centering
\begin{minipage}[t]{0.49\linewidth}\centering
\caption{Performance of hair segmentation on the test set.}
\label{tbl1:comparison_seg_result}
\resizebox{0.95\linewidth}{!}{
\begin{tabular}{c|c|ccc}
\toprule
\multicolumn{2}{c|}{Approach} & Pixel-F1 & Jaccard & Dice  \\ 
\midrule
\multicolumn{2}{c|}{Shih \textit{\scriptsize et al.}~\cite{shih2014:unsupervised}} & $0.706$   & $0.348$   & $0.512$ \\
\multicolumn{2}{c|}{Yue \textit{\scriptsize et al.}~\cite{yue2021:hair}}           & $0.794$  & $0.493$ & $0.654$ \\
\multicolumn{2}{c|}{Kim \textit{\scriptsize et al.}~\cite{kim2017:evaluation}}     & $0.815$  & $0.561$  & $0.708$ \\
\multicolumn{2}{c|}{SAM~\cite{kirillov2023:segment}}                   &  $0.503$  & $0.361$  & $0.502$ \\ 
\midrule
\multirow{3}{*}{Ours}
& $\hat{M}$           &  $0.853$  & $0.604$  & $0.748$ \\ 
& $M_\text{AP}$       &  $0.836$  & $0.595$  & $0.743$ \\ 
& $M$                 & $\textbf{0.868}$  & $\textbf{0.649}$  &  $\textbf{0.786}$  \\
\bottomrule
\end{tabular}
}
\end{minipage}\hfill%
\begin{minipage}[t]{0.49\linewidth}\centering
% \vspace{0.45cm} 
\caption{Quantitative analysis of image-to-image translation.}
\label{tab:quan_diffitm}
\resizebox{0.7\linewidth}{!}{
\begin{tabular}{@{}c|cc@{}}
\toprule
Model    & FID ($\downarrow$) & LPIPS ($\downarrow$) \\ \midrule
DiffuseIT &$138.42$&$0.463$\\
AGG       &$141.70$&$0.492$  \\
Ours      &$\textbf{74.84}$&$\textbf{0.353}$\\ \bottomrule
\end{tabular}
}
\end{minipage}
\end{table}

\subsection{Hair Segmentation}
\label{sec:exp_seg}

Because narrow hairs are difficult to segment, many existing unsupervised methods still rely on traditional computer vision for hair-specific challenges. 
Accordingly, we compare prior scalp segmentation approaches~\cite{shih2014:unsupervised,yue2021:hair,kim2017:evaluation} as baselines and also evaluate the foundational model SAM~\cite{kirillov2023:segment}. 
In addition, we conduct an ablation study on our final mask $M$ and its intermediate versions $\hat{M}$ and $M_{\mathrm{AP}}$.

\noindent\textbf{Quantitative results.}
Table~\ref{tbl1:comparison_seg_result} reveals that our methods surpass the performance of existing hair segmentation techniques.
In particular, the approach of combining the advantages of the two masks, $\hat{M}$ and $M_\text{AP}$ using the logical AND operator in $M$ showed the best performance.
These results show the limitations of traditional computer vision techniques used in previous studies for image segmentation, revealing a lack of understanding in capturing the intricate patterns of hair and the scalp.
Additionally, SAM was less effective for automatic segmentation when used without specific guidance.

\noindent\textbf{Qualitative results.}
As shown in Figure~\ref{fig:comp_seg}, our approach demonstrates effective hair segmentation with robustness to noise, providing clear and accurate hair segmentation compared to previous methods.
Furthermore, it shows that $\hat{M}$ faces challenges in clearly capturing hair, and it exhibits robustness against noise such as dandruff.
Conversely, $M_\text{AP}$ captures the hair well but is less robust to noise.
Therefore, the combination of the two masks, $M$, demonstrates the mitigation of the drawbacks of each mask.

\subsection{Synthetic Image Generation}
\label{sec:scalp_cls}
For the evaluation of \emph{DiffuseIT-M}, we compared our model against DiffuseIT~\cite{kwon2022:diffusion} and AGG~\cite{kwon2023:improving} as baselines for the image-to-image translation model.
This evaluation demonstrates that our model not only achieves high fidelity in image-to-image translation but also effectively preserves the desired hair details.

\begin{figure}[t]
    \centering
    \begin{minipage}[t]{0.49\linewidth}
        \centering
        \includegraphics[width=\linewidth]{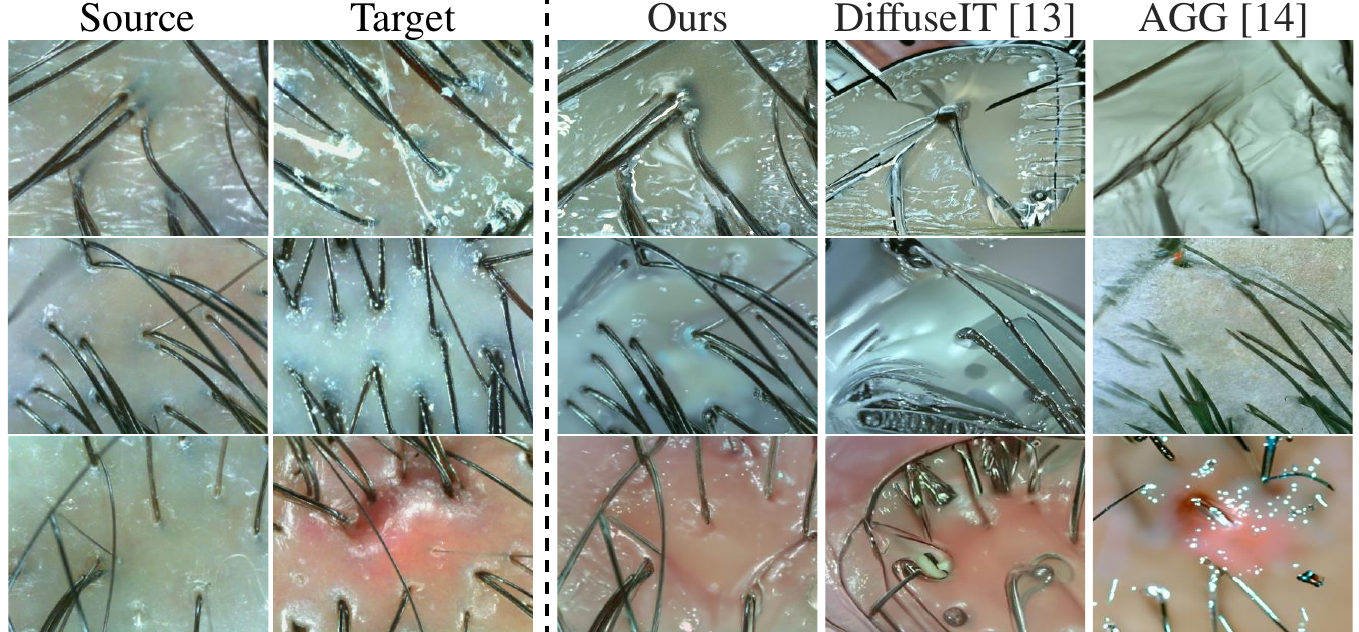}
        \caption{
        Image translation results with different generative models, where the goal is to preserve source hairlines while changing the scalp.
        }
        \label{fig:comp_gen}
    \end{minipage}
    \hfill
    \begin{minipage}[t]{0.49\linewidth}
        \centering
        \includegraphics[width=\linewidth]{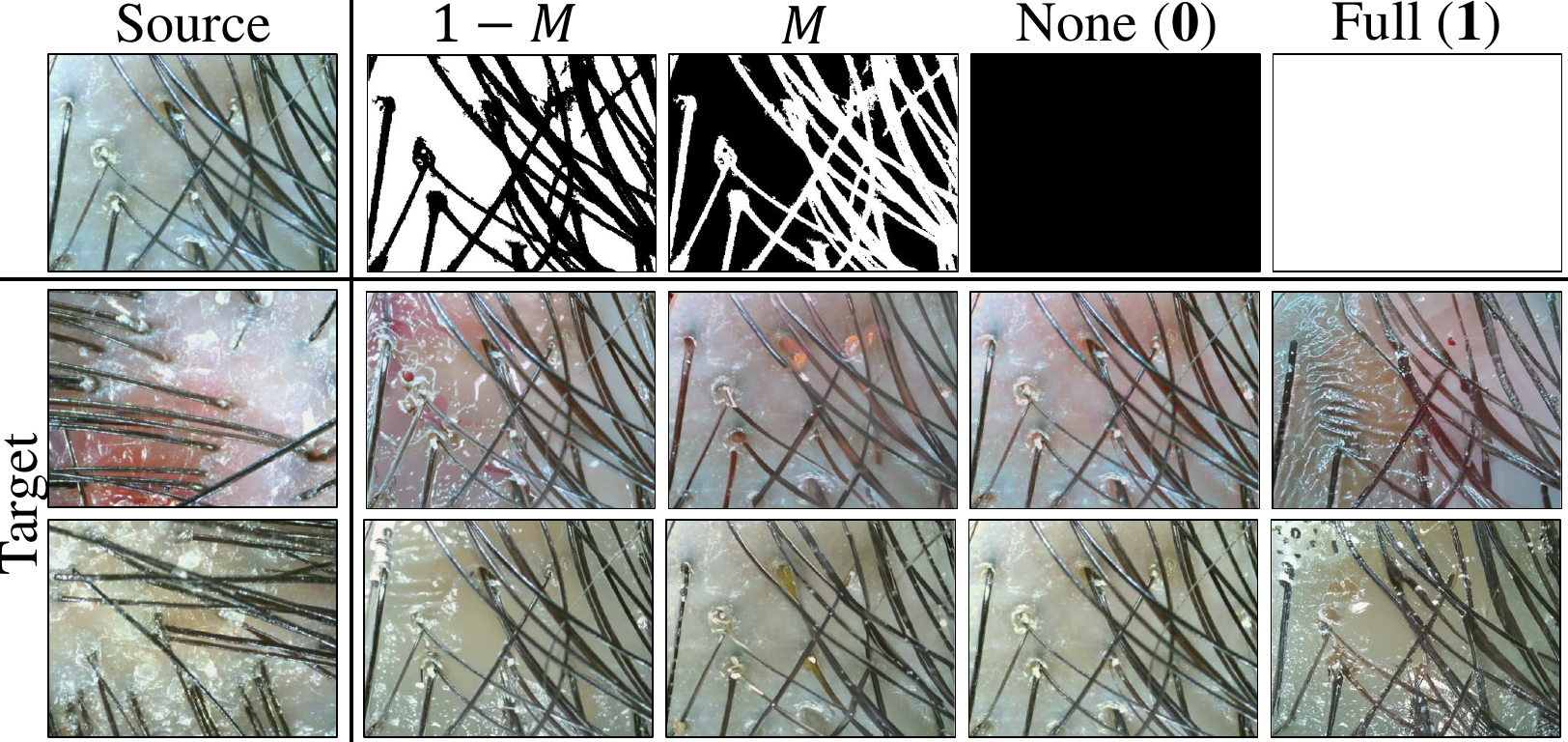}
        \caption{
        Image translation results using various mask guidance.
        Note that our approach is guided by $1 - M$.
        }
        \label{fig:abl_gen}
    \end{minipage}
\end{figure}

\noindent \textbf{Quantitative results.}
We have selected to employ the FID~\cite{heusel2017:gans} and LPIPS~\cite{zhang2018:unreasonable} scores for fidelity evaluation using images from our augmentation dataset, with DiffuseIT and AGG serving as baseline models.
Table~\ref{tab:quan_diffitm} reveals that \emph{DiffuseIT-M} outperforms other models in both metrics, indicating superior image fidelity. 
This high-quality image generation is attributed to our model's effective implementation of mask guidance.

\noindent \textbf{Qualitative results.}
Figure~\ref{fig:comp_gen} shows that both DiffuseIT and AGG models fail to preserve the hair content information from the source image.
Furthermore, these models tended to compromise overall information and were unable to transfer the semantic information.
However, our model successfully preserved hair content information and transferred the semantic information.

\noindent \textbf{Effect of mask guidance.}
We conducted experiments to examine the impact of mask guidance on hair information preservation during image translation. 
As illustrated in Figure~\ref{fig:abl_gen}, our method, guided by the mask $1 - M$, effectively retains hair features while successfully transferring the semantic attributes of the target image onto the scalp. 
In contrast, using the reverse mask, $M$, leads to only minor alterations in scalp color from the target image, with a notable transfer of hair semantic information from the target.
When no mask ($\mathbf{0}$) is applied, the translation results in minimal color change, failing to transfer conditions like dandruff from the target. 
Conversely, with a full mask ($\mathbf{1}$), both hair and scalp features are subjected to changes. 
This differentiation in results highlights the importance of mask guidance in preserving specific image features, demonstrating the versatility of our approach in handling different translation objectives.

%---------------------------------------------------------------------------------------------------------------------------------------
\begin{table*}[t!]
\caption{
Performance of scalp condition classification with various augmentation methods, denoted after ``+'' symbol, on the test set.
The second column displays the overall macro-F1 score, while the columns from the third onward show the F1 scores for each severity level of the three diseases.
}
\resizebox{\textwidth}{!}{%
\begin{tabular}{@{}l||c|cccc|cccc|cccc@{}}
\toprule
\multirow{2}{*}{Model} & F1    & \multicolumn{4}{c|}{Dandruff}      & \multicolumn{4}{c|}{Sebum}         & \multicolumn{4}{c}{Erythema}      \\
                       & macro & good  & mild  & moderate & severe & good  & mild  & moderate & severe & good  & mild  & moderate & severe \\ \cmidrule(r){1-1}   \cmidrule(l){2-14} 

DenseNet
&   $0.582$        & $0.796$  & $0.514$      &  $0.592$        &  $0.614$   & $\textbf{0.554}$    & $0.601$   & $0.641$     &  $0.000$    &  $\textbf{0.776}$ & $0.729$ &    $0.565$  &  $0.601$  \\     
\hspace{0.1cm}+ Gaussian Noise  &   $0.567$   &  $0.780$   &  $0.497$   &  $0.566$   & $0.597$     & $0.471$    &$0.581$     &  $0.595$    & $0.000$   &$0.751$&$0.712$&$0.614$&$0.635$\\      
\hspace{0.1cm}+ AugMix
&$0.525$&$0.789$&$0.501$&$0.589$&$0.000$&$0.504$&$0.588$&$0.634$&$0.000$&$0.743$&$0.718$&$0.596$&$0.585$\\ 

\hspace{0.1cm}+ DiffuseIT
&$0.608$&$0.809$&$0.482$&$0.604$&$0.650$&$0.536$&$0.613$&$0.625$&$0.202$&$0.774$&$0.740$&$0.621$&$0.639$\\  
\hspace{0.1cm}+ AGG
&$0.610$&$0.811$&$0.480$&$0.591$&$0.654$&$0.518$&$0.598$&$0.612$&$0.300$&$0.771$&$\textbf{0.740}$&$\textbf{0.629}$&$0.621$\\  
\hspace{0.1cm}+ Ours&$\textbf{0.636}$&$\textbf{0.820}$&$\textbf{0.541}$&$\textbf{0.625}$&$\textbf{0.665}$&$0.536$&$\textbf{0.617}$&$\textbf{0.641}$&$\textbf{0.430}$&$0.758$&$0.734$&$0.621$&$\textbf{0.641}$\\

\midrule 

EfficientFormerV2
& $0.569$ & $0.795$ & $0.417$ & $0.598$ &$0.628$&$0.526$&$0.565$ & $0.628$ &$ 0.000$& $0.772$ &$0.709$&$0.623$&$0.569$\\
\hspace{0.1cm}+ Gaussian Noise &  $0.562$ & $0.780$ & $0.477$ & $0.566$       & $0.633$ & $0.460$   & $0.585$  & $0.550$ & $0.000$   &$0.742$&$0.714$&$0.598$&$0.637$\\
\hspace{0.1cm}+ AugMix
&$0.577$&$0.789$&$0.494$&$0.592$&$0.635$&$0.519$&$0.593$&$0.623$&$0.000$&$0.746$&$0.724$&$0.620$&$0.590$\\

\hspace{0.1cm}+ DiffuseIT
&$0.596$ &$0.798$&$0.441$&$0.598$&$0.632$&$0.526$&$0.595$&$0.606$&$0.236$&$0.766$&$0.715$&$0.612$&$0.621$\\
\hspace{0.1cm}+ AGG
&$0.610$&$0.801$&$0.509$&$0.604$&$0.626$&$0.511$&$0.583$&$0.608$&$0.300$&$\textbf{0.787}$&$0.736$&$0.624$&$0.628$\\  
\hspace{0.1cm}+ Ours            
&$\textbf{0.635}$&	$\textbf{0.807}$&	$\textbf{0.529}$	&$\textbf{0.619}$&	$\textbf{0.669}$&	$\textbf{0.535}$&	$\textbf{0.613}$	&$\textbf{0.632}$&$\textbf{0.406}$	&$0.781$	&$\textbf{0.738}$	&$\textbf{0.639}$&	$\textbf{0.648}$ \\

\bottomrule
\end{tabular}
}

\label{tab:cls_result_bce}
\end{table*}
%---------------------------------------------------------------------------------------------------------------------------------------

\subsection{Scalp Condition Classification}
To demonstrate the effectiveness of our augmentation method using generated images, we employed two different models as the classification backbone: DenseNet ~\cite{huang2017:densely} as a CNN and EfficientFormerV2~\cite{li2023:rethinking} as a Transformer.
As summarized in Table~\ref{tab:cls_result_bce}, we compared only those augmentation methods that preserve a one-to-one correspondence between each image and its original, unambiguous set of condition labels -- Gaussian noise, AugMix~\cite{hendrycks2019:augmix}, DiffuseIT~\cite{kwon2022:diffusion}, and AGG~\cite{kwon2023:improving}.
Our approach, which specifically employs \emph{DiffuseIT-M}, achieved the highest performance in both models. 
Notably, classifying the \textit{severe} sebum class proved to be especially challenging when using non-generative augmentation methods. 
This difficulty arises primarily due to the extreme scarcity of samples for this class.
The augmentation of the training dataset with generative models led to enhanced performance compared to the baseline.
Our model, in particular, exhibited superior accuracy compared to DiffuseIT and AGG, which struggled to preserve the essential information of the hair effectively. 
This underscores the significance of incorporating both the scalp style details and the hair content information in the scalp disorder classification.

\section{Conclusion and Discussion}
\label{sec:conclusion}
In this work, we introduced \textsc{ScalpVision}, a diagnostic system designed for a complete evaluation of scalp health.
Our approach combines label-free hair segmentation -- based on a naive segmentation model and a foundation segmentation model (SAM) -- with diffusion-based data augmentation to address data imbalance and preserve critical hair features.
However, scalp disorders are affected by both the condition of the scalp and hair characteristics. Therefore, we plan to incorporate hair information to broaden our research to conditions such as alopecia, beyond the three scalp diseases. We see \textsc{ScalpVision} as an important step toward a more general diagnostic system for dermatological applications.

\begin{credits}
\subsubsection{\ackname} This work was supported by the National Research Foundation of Korea(NRF) grant funded by the Korea government(MSIT) (No. RS-2024-00354218 and RS-2024-00353125). Junhyug Noh was supported by Institute of Information \& communications Technology Planning \& Evaluation (IITP) grant funded by the Korea government (MSIT) (No.RS-2022-00155966).
% \subsubsection{\discintname}
% The authors have no competing interests to declare that are relevant to the content of this article.
\end{credits}

\bibliographystyle{splncs04}
\bibliography{main}

\begin{thebibliography}{10}
\providecommand{\url}[1]{\texttt{#1}}
\providecommand{\urlprefix}{URL }
\providecommand{\doi}[1]{https://doi.org/#1}

\bibitem{aihub:2020}
{AI Hub}: {Scalp and Hair Follicle Image Dataset} (2020), \url{https://aihub.or.kr/aihubdata/data/view.do?dataSetSn=207}, [Online; accessed 25 February 2025]

\bibitem{avrahami2022:blended}
Avrahami, O., Lischinski, D., Fried, O.: Blended diffusion for text-driven editing of natural images. In: Proceedings of the IEEE/CVF Conference on Computer Vision and Pattern Recognition. pp. 18208--18218 (2022)

\bibitem{borda2015:seborrheic}
Borda, L.J., Wikramanayake, T.C.: Seborrheic dermatitis and dandruff: a comprehensive review. Journal of clinical and investigative dermatology  \textbf{3}(2) (2015)

\bibitem{caron2021:emerging}
Caron, M., Touvron, H., Misra, I., J{\'e}gou, H., Mairal, J., Bojanowski, P., Joulin, A.: Emerging properties in self-supervised vision transformers. In: Proceedings of the IEEE/CVF international conference on computer vision. pp. 9650--9660 (2021)

\bibitem{chang2020:scalpeye}
Chang, W.J., Chen, L.B., Chen, M.C., Chiu, Y.C., Lin, J.Y.: Scalpeye: A deep learning-based scalp hair inspection and diagnosis system for scalp health. IEEE Access  \textbf{8},  134826--134837 (2020)

\bibitem{crowson2022:vqgan}
Crowson, K., Biderman, S., Kornis, D., Stander, D., Hallahan, E., Castricato, L., Raff, E.: Vqgan-clip: Open domain image generation and editing with natural language guidance. In: European Conference on Computer Vision. pp. 88--105. Springer (2022)

\bibitem{dosovitskiy2021:image}
Dosovitskiy, A., Beyer, L., Kolesnikov, A., Weissenborn, D., Zhai, X., Unterthiner, T., Dehghani, M., Minderer, M., Heigold, G., Gelly, S., Uszkoreit, J., Houlsby, N.: An image is worth 16x16 words: Transformers for image recognition at scale (2021)

\bibitem{elewski2005clinical}
Elewski, B.E.: Clinical diagnosis of common scalp disorders. In: Journal of Investigative Dermatology Symposium Proceedings. vol.~10, pp. 190--193. Elsevier (2005)

\bibitem{he2016:deep}
He, K., Zhang, X., Ren, S., Sun, J.: Deep residual learning for image recognition. In: Proceedings of the IEEE conference on computer vision and pattern recognition. pp. 770--778 (2016)

\bibitem{hendrycks2019:augmix}
Hendrycks, D., Mu, N., Cubuk, E.D., Zoph, B., Gilmer, J., Lakshminarayanan, B.: Augmix: A simple data processing method to improve robustness and uncertainty. arXiv preprint arXiv:1912.02781  (2019)

\bibitem{heusel2017:gans}
Heusel, M., Ramsauer, H., Unterthiner, T., Nessler, B., Hochreiter, S.: Gans trained by a two time-scale update rule converge to a local nash equilibrium. Advances in neural information processing systems  \textbf{30} (2017)

\bibitem{huang2017:densely}
Huang, G., Liu, Z., Van Der~Maaten, L., Weinberger, K.Q.: Densely connected convolutional networks. In: Proceedings of the IEEE conference on computer vision and pattern recognition. pp. 4700--4708 (2017)

\bibitem{kim2017:evaluation}
Kim, H., Kim, W., Rew, J., Rho, S., Park, J., Hwang, E.: Evaluation of hair and scalp condition based on microscopy image analysis. In: 2017 International conference on platform technology and service (PlatCon). pp.~1--4. IEEE (2017)

\bibitem{kim2022:hair}
Kim, J.H., Kwon, S., Fu, J., Park, J.H.: Hair follicle classification and hair loss severity estimation using mask r-cnn. Journal of Imaging  \textbf{8}(10), ~283 (2022)

\bibitem{kirillov2023:segment}
Kirillov, A., Mintun, E., Ravi, N., Mao, H., Rolland, C., Gustafson, L., Xiao, T., Whitehead, S., Berg, A.C., Lo, W.Y., et~al.: Segment anything. arXiv preprint arXiv:2304.02643  (2023)

\bibitem{kwon2022:diffusion}
Kwon, G., Ye, J.C.: Diffusion-based image translation using disentangled style and content representation. In: The Eleventh International Conference on Learning Representations (2022)

\bibitem{kwon2023:improving}
Kwon, G., Ye, J.C.: Improving diffusion-based image translation using asymmetric gradient guidance. arXiv preprint arXiv:2306.04396  (2023)

\bibitem{li2023:rethinking}
Li, Y., Hu, J., Wen, Y., Evangelidis, G., Salahi, K., Wang, Y., Tulyakov, S., Ren, J.: Rethinking vision transformers for mobilenet size and speed. In: Proceedings of the IEEE/CVF International Conference on Computer Vision. pp. 16889--16900 (2023)

\bibitem{li2022:efficientformer}
Li, Y., Yuan, G., Wen, Y., Hu, J., Evangelidis, G., Tulyakov, S., Wang, Y., Ren, J.: Efficientformer: Vision transformers at mobilenet speed. Advances in Neural Information Processing Systems  \textbf{35},  12934--12949 (2022)

\bibitem{loshchilov2016:sgdr}
Loshchilov, I., Hutter, F.: Sgdr: Stochastic gradient descent with warm restarts. arXiv preprint arXiv:1608.03983  (2016)

\bibitem{loshchilov2017:decoupled}
Loshchilov, I., Hutter, F.: Decoupled weight decay regularization. arXiv preprint arXiv:1711.05101  (2017)

\bibitem{otsu1979:threshold}
Otsu, N.: A threshold selection method from gray-level histograms. IEEE transactions on systems, man, and cybernetics  \textbf{9}(1),  62--66 (1979)

\bibitem{panjwani2009:early}
Panjwani, S.: Early diagnosis and treatment of discoid lupus erythematosus. The Journal of the American Board of Family Medicine  \textbf{22}(2),  206--213 (2009)

\bibitem{pratt2017:alopecia}
Pratt, C.H., King, L.E., Messenger, A.G., Christiano, A.M., Sundberg, J.P.: Alopecia areata. Nature reviews Disease primers  \textbf{3}(1),  1--17 (2017)

\bibitem{qin2020:u2}
Qin, X., Zhang, Z., Huang, C., Dehghan, M., Zaiane, O.R., Jagersand, M.: U2-net: Going deeper with nested u-structure for salient object detection. Pattern recognition  \textbf{106},  107404 (2020)

\bibitem{sakuma2012:oily}
Sakuma, T.H., Maibach, H.I.: Oily skin: an overview. Skin pharmacology and physiology  \textbf{25}(5),  227--235 (2012)

\bibitem{seo2020:trichoscopy}
Seo, S., Park, J.: Trichoscopy of alopecia areata: hair loss feature extraction and computation using grid line selection and eigenvalue. Computational and Mathematical Methods in Medicine  \textbf{2020} (2020)

\bibitem{shih2014:unsupervised}
Shih, H.C.: An unsupervised hair segmentation and counting system in microscopy images. IEEE Sensors Journal  \textbf{15}(6),  3565--3572 (2014)

\bibitem{waskiel2023:differential}
Wa{\'s}kiel-Burnat, A., Czuwara, J., Blicharz, L., Olszewska, M., Rudnicka, L.: Differential diagnosis of red scalp. the importance of trichoscopy. Clinical and Experimental Dermatology p. llad366 (2023)

\bibitem{rw2019:timm}
Wightman, R.: Pytorch image models. \url{https://github.com/rwightman/pytorch-image-models} (2019). \doi{10.5281/zenodo.4414861}

\bibitem{xie2017:aggregated}
Xie, S., Girshick, R., Doll{\'a}r, P., Tu, Z., He, K.: Aggregated residual transformations for deep neural networks. In: Proceedings of the IEEE conference on computer vision and pattern recognition. pp. 1492--1500 (2017)

\bibitem{yue2021:hair}
Yue, G., Ji, C., Yang, Y., et~al.: Hair counting method based on image processing technology. Journal of Artificial Intelligence Practice  \textbf{4}(1),  23--29 (2021)

\bibitem{zhang2018:unreasonable}
Zhang, R., Isola, P., Efros, A.A., Shechtman, E., Wang, O.: The unreasonable effectiveness of deep features as a perceptual metric. In: Proceedings of the IEEE conference on computer vision and pattern recognition. pp. 586--595 (2018)

\bibitem{Zhang1984:morphSkel}
Zhang, T.Y., Suen, C.Y.: A fast parallel algorithm for thinning digital patterns. Commun. ACM  \textbf{27}(3),  236–239 (mar 1984). \doi{10.1145/357994.358023}, \url{https://doi.org/10.1145/357994.358023}

\end{thebibliography}
\clearpage
\appendix
\section{Detailed Overview of Scalp Diseases}
The dataset from AI-Hub\footnote{https://aihub.or.kr} categorizes scalp images into three primary conditions: \textit{dandruff}, \textit{excess sebum}, and \textit{erythema}.

Dandruff, also referred to as a milder manifestation of seborrheic dermatitis, is characterized by the non-inflammatory exfoliation of dead epidermal cells from the scalp. While it can induce mild itching, it generally does not precipitate erythema or the formation of scabs~\cite{borda2015:seborrheic}.

Hyperseborrhea, the excessive production of sebum, represents a common aesthetic concern, manifested through the secretion of excess oil from hypertrophic sebaceous glands. This condition results in a shiny, oily skin appearance. Although sebum plays a crucial role in maintaining skin hydration and its protective barrier, its excessive secretion can lead to various dermatological issues. One such issue is the formation of sebum plugs, which are small, yellowish, or pale bumps that appear on the skin~\cite{sakuma2012:oily}.

Scalp erythema, also known as red scalp, is characterized by widespread redness across the scalp. It can arise from several conditions, including psoriasis, seborrheic dermatitis, contact dermatitis, diffuse lichen planopilaris, dermatomyositis, and scalp rosacea~\cite{waskiel2023:differential}.

\section{Implementation Details}
\label{sec:appendix_details}
\noindent \textbf{Hair segmentation. }For our heuristic-driven hair segmentation, we utilized the U$^{2}$-Net~\cite{qin2020:u2} official source code to train $3{,}000$ pseudo image-label pairs. 
The training parameters included a batch size of $32$, a constant learning rate of $0.001$, $100$ training epochs, and the Adam optimizer. 
The output from U$^{2}$-Net was binarized to obtain $\hat{M}$, using a threshold of $0.5$.

Since other previous studies, excluding SAM~\cite{kirillov2023:segment}, do not have an official codebase, we implemented their approaches based on descriptions in their papers, using OpenCV.
For instance, \cite{shih2014:unsupervised} used contrast stretching and binary thresholding to derive the hair mask. 
\cite{yue2021:hair} applied morphological operations, while \cite{kim2017:evaluation} used Otsu's method~\cite{otsu1979:threshold} for hair mask acquisition. 
In evaluating SAM, the mask with the highest Intersection over Union (IoU) score was selected as the final prediction.

\noindent \textbf{Image augmentation.}
For \emph{DiffuseIT-M}, we construct a loss function for image translation while preserving hair content. This function is defined as follows:
\begin{equation}
    \begin{gathered}
    \ell_{total}(x; x_{src}, x_{trg}, M) = \lambda_{1}\ell_{style} + \lambda_{2}\ell_{content} \\+ \lambda_{3}\ell_{mask} + \lambda_{4}\ell_{sem} + \lambda_{5}\ell_{rng}.
    \end{gathered}
    \label{eq:diffit_loss}
\end{equation}

To incorporate the semantic information of the target image, we establish the style loss function, $\ell_{sty}$. 
This function leverages the \texttt{[CLS]} token from the last layer of DINO-ViT~\cite{caron2021:emerging}. 
Denoting the \texttt{[CLS]} tokens as $\mathbf{c}$, the style loss is expressed as:
\begin{equation}
    \begin{gathered}
        \ell_{sty}(x_{trg}, \hat{x}_0 (x_t)) = ||\mathbf{c}(x_{trg}) - \mathbf{c}(\bar{x})||_{2} \\+ \lambda_{mse} || x_{trg} - \hat{x}_0 (x_t)||_{2},
    \end{gathered}
\end{equation}
where $\lambda_{mse}$ is set to $3{,}000$, and the weight for $\ell_{style}$, $\lambda_{1}$, is set to $2{,}000$.
%------------------------------------------------------------------------
\begin{figure}[t!]
    \centering
    \includegraphics[width=0.8\linewidth]{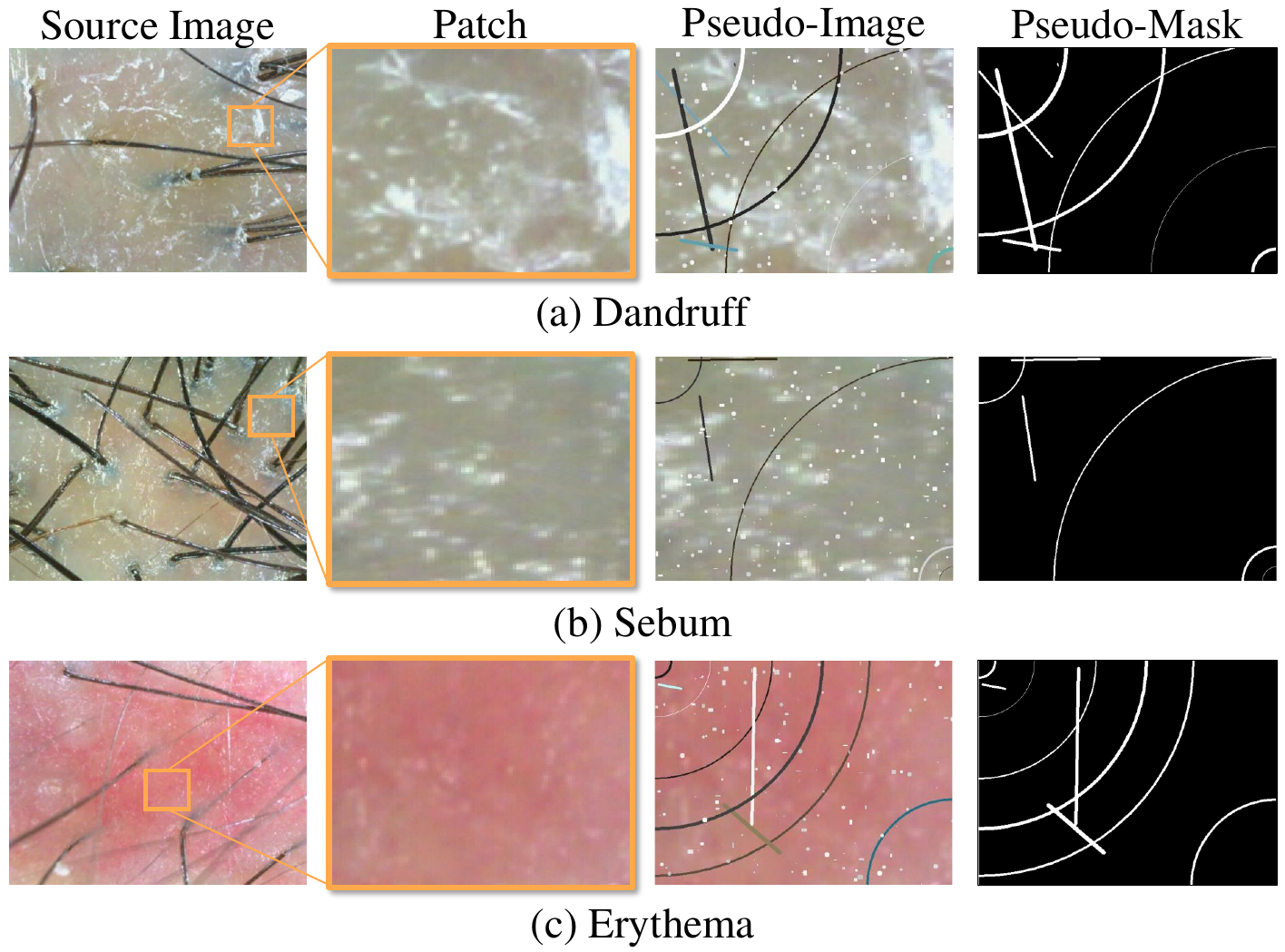}
    \caption{Examples of pseudo images and their corresponding masks for hair segmentation.}
    \label{fig:sppl_psuedo}
\end{figure}
%------------------------------------------------------------------------
The content loss, $\ell_{content}$, is designed to preserve the structure of source images. 
Let $k^{l}_{i}(x)$ represent the $i$-th key extracted from the $l$-th multi-head self-attention layer in DINO-ViT for image $x$. 
The content loss is then defined as:
\begin{equation}
    \ell_{content} = \lambda_{sim}\ell_{sim}(x_{src}, \hat{x}_0 (x_t)) + \lambda_{con}\ell_{con}(x_{src}, \hat{x}_0 (x_t)),
\end{equation}
where the similarity loss, $\ell_{sim}$, and the content loss, $\ell_{con}$, are 
\begin{align}
\ell_{sim}(x_{src}, \hat{x}_0 (x_t)) =  || \text{cos}_{ij}(x_{src}), \text{cos}_{ij}(\hat{x}_0 (x_t)) ||_{2}, \\
\ell_{con}(x_{src}, \hat{x}_0 (x_t)) = \text{infoNCE}(k^{l}_{i}(x_{src}), k^{l}_{i}(\hat{x}_0 (x_t))),
\end{align}
with $cos_{ij}(x)$ representing the cosine distance between $k^{l}_{i}(x)$ and $k^{l}_{j}(x)$.
The weights $\lambda_{sim}$ and $\lambda_{con}$ are set to $1{,}000$ and $200$, respectively. 
Additionally, weights $\lambda_{3}$, $\lambda_{4}$, and $\lambda_{5}$ are set to $1,000$, $100$, and $200$, respectively.
During our experiment, the model is configured to generate images with a resolution of $256 \times 256$ pixels, utilizing a denoising step of $1{,}000$.

For the implementation of previous studies, we used the official source code of DiffuseIT~\cite{kwon2022:diffusion} and AGG~\cite{kwon2023:improving}, conducting experiments with their provided hyperparameters. 
Non-generative image augmentations were performed using AugMix~\cite{hendrycks2019:augmix} through the torchvision library and Gaussian noise augmentation via the PyTorch library.

\noindent \textbf{Scalp disease and severity diagnosis.}
\label{suppl:hyperparaemter_classification}
For the classification task, we fine-tuned two models: DenseNet169~\cite{huang2017:densely} (CNN-based) and EfficientFormerV2~\cite{li2023:rethinking} (Transformer-based), using pre-trained weights from the timm library~\cite{rw2019:timm}. 
Fine-tuning involved a batch size of $128$, a learning rate of $0.0001$ with a CosineAnnealingWarmRestarts scheduler~\cite{loshchilov2016:sgdr}, $50$ training epochs, and the AdamW optimizer~\cite{loshchilov2017:decoupled}.

\section{Pseudo Image and Mask Visualization}
\label{sec:suppl_seg}
To create a diverse pseudo training set, we extracted scalp patch images from areas without hair in nine different scalp images. 
Each image represented a unique disease at a specific severity level. 
As illustrated in Figure~\ref{fig:sppl_psuedo}, we introduced a variety of hair types by inserting straight and curved lines in blue, brown, black, and white colors, each with differing thicknesses. 
To simulate common scalp noise, such as dandruff, white circular elements were added to the pseudo images. 
Our codebase contains further details on this process.

\section{Scalp Disease and Severity Classification}
\label{sec:suppl_scalp_class}
This section outlines our data augmentation approach for classifying scalp diseases and their severities and presents additional experimental results.
%-------------------------------------------------------------------------
\begin{algorithm}[t!]
\caption{Calculation of sampling ratios for each disease}
\label{algorithm:sampling_ratio}
\small
\textbf{Input}: $severities$: A collection of records for each severity. Assume that each element of $severities$ is classified as 0 (\textit{good}), 1 (\textit{mild}), 2 (\textit{moderate}), and 3 (\textit{severe}).

\textbf{Output}: $ratios$: Sampling ratio among four severity levels.

\begin{algorithmic}[1]
\STATE $\epsilon \leftarrow 1 \times 10^{-9}$

\STATE $sevCounts$ $\leftarrow$ $\{ \}$
\FOR{each $sevLevel$ in $[0,1,2,3]$}
    \STATE $sevCounts[sevLevel]$ $\leftarrow$ $0$
\ENDFOR

\FOR{each $severity$ in $severities$}
    \STATE $ sevCounts[severity]$ $\leftarrow$ $sevCounts[severity] + 1$
    
\ENDFOR
\STATE $invCounts$ $\leftarrow$ $\{\}$

\FOR{each $sevLevel$ in $[0,1,2,3]$}
    \STATE $invCounts[sevLevel]$ $\leftarrow$ $1/(sevCounts[sevLevel]+\epsilon)$
\ENDFOR

\STATE $normFactor$ $\leftarrow$ $\sum_{sevLevel=0}^{3} invCounts[sevLevel]$

\STATE $ratios$ $\leftarrow$ $[]$
\FOR{each $sevLevel$ in $[0,1,2,3]$}
    \STATE $ratio \gets invCounts[sevLevel] / normFactor$
    \STATE $ratios[sevLevel]$ $\leftarrow$ $ratio$
\ENDFOR
    \STATE \textbf{return} $ratios$
\end{algorithmic}        
\normalsize
\end{algorithm}

% --------------------------------
\begin{figure}[t]
    \centering
    \includegraphics[width=\linewidth]{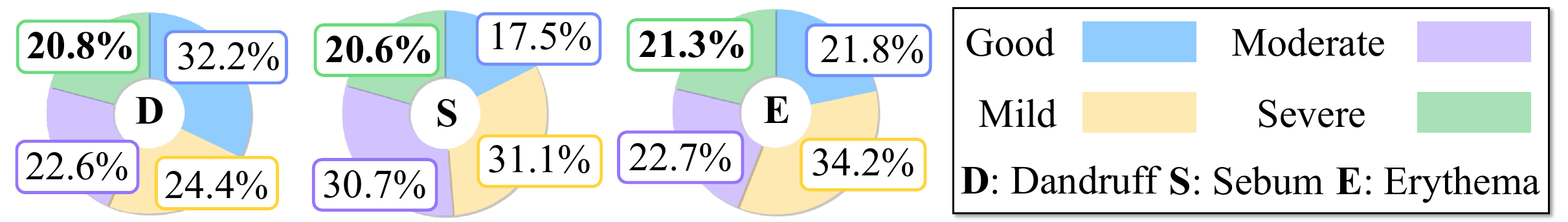}
    \caption{Distribution of data augmented using \emph{DiffuseIT-M}. For comparison, the original data distribution is presented in Figure~\ref{fig:data_stat}.}
    \label{fig:data_distribution_after_aug}
\end{figure}
%-------------------------------------------------------------------------
%-------------------------------------------------------------------------
\subsection{Data Augmentation Strategy}
Addressing the issue of data imbalance in the dataset, we implemented a strategy to translate randomly selected images into classes with fewer samples. 
We used random selection for the source images and weighted sampling for target images, where the likelihood of choosing an image was inversely proportional to the number of samples in its severity class. 
This method favored the selection of underrepresented classes. 
The algorithm to calculate these sampling weights is detailed in Algorithm~\ref{algorithm:sampling_ratio}.

Figure~\ref{fig:data_distribution_after_aug} displays the data distribution after this augmentation. 
The post-augmentation distribution shows a more balanced representation across various classes, especially a rise in the \textit{severe} category and a reduction in the \textit{good} category. 
This balanced dataset played a crucial role in enhancing our model's performance by providing an even distribution of the training samples. 
The same augmentation strategy was applied across all diffusion-based methods.
%------------------------------------------------------------------
\begin{table}[t]
\centering
\caption{
Quantitative comparison of classification performance across different backbone models. ``Baseline'' refers to performance without any augmentation methods, while results following the ``+'' symbol indicate the use of various augmentation methods. Values in the table represent the macro-F1 scores.
}
\resizebox{0.7\columnwidth}{!}{%
% \begin{tabular}{@{}ccccc@{}}
\begin{tabular}{l|cc|cc}
\toprule
\multirow{2}{*}{Method}              & \multicolumn{2}{c|}{CNN}                             & \multicolumn{2}{c}{Transformer}   \\ \cmidrule(l){2-5} 
                               & ResNet        & \multicolumn{1}{c|}{ResNeXt} & EfficientFormer & ViT             \\ \midrule
\multicolumn{1}{l|}{Baseline}  & $0.5709$          & \multicolumn{1}{c|}{$0.6186$}             & $0.5300$            & $0.4934$          \\
\multicolumn{1}{l|}{+ Gaussian Noise}    & $0.4304$          & \multicolumn{1}{c|}{$0.5663$}             & $0.5742$          & $0.4306$          \\
\multicolumn{1}{l|}{+ AugMix}    & $0.4186$          & \multicolumn{1}{c|}{$0.5247$}             & $0.5348$          & $0.4747$          \\
\multicolumn{1}{l|}{+ DiffuseIT} & $0.6064$          & \multicolumn{1}{c|}{$0.6107$}             & $0.6120$           & $0.5767$          \\
\multicolumn{1}{l|}{+ AGG}       & $0.5978$          & \multicolumn{1}{c|}{$0.6251$}             & $0.6104$            & $0.5771$          \\
\multicolumn{1}{l|}{+ Ours}      & $\textbf{0.6128}$ & \multicolumn{1}{c|}{$\textbf{0.6292}$}             & $\textbf{0.6170}$  & $\textbf{0.5861}$ \\ \bottomrule
\end{tabular}%
}

\label{tab:sppl_quan_cls_backbone}
\end{table}
%------------------------------------------------------------------
\begin{figure*}[ht!]
    \centering
    \includegraphics[width=0.95\columnwidth]{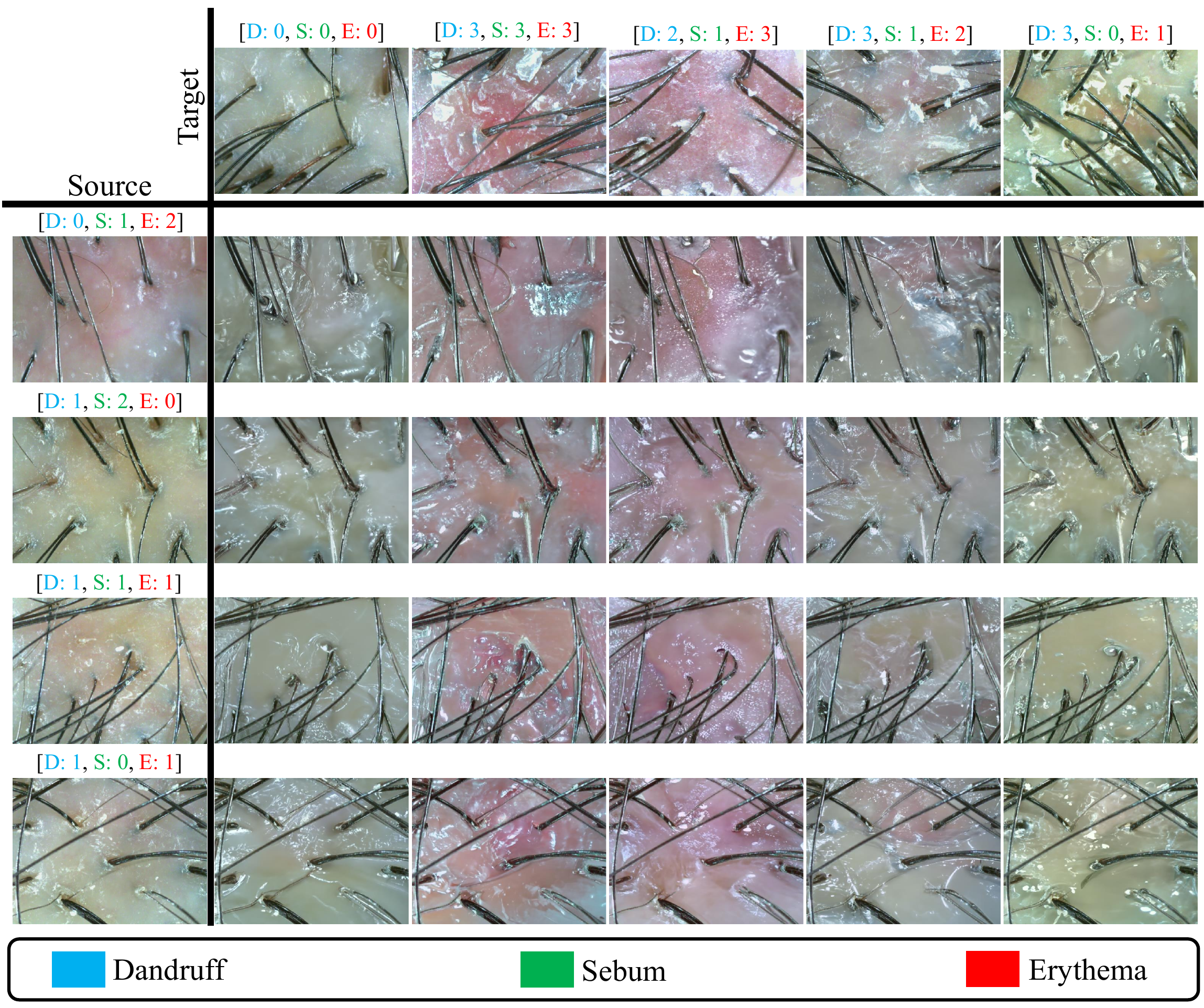}
    \caption{
    Qualitative results of \emph{DiffuseIT-M}. 
    This figure illustrates the results for various scalp disease conditions, with severity levels indicated as 0 (\textit{good}), 1 (\textit{mild}), 2 (\textit{moderate}), and 3 (\textit{severe}). Scalp diseases are color-coded for clarity: \textcolor{blue}{blue} represents dandruff, \textcolor{green}{green} signifies excess sebum, and \textcolor{red}{red} denotes erythema.
    }
    \label{fig:sppl_diffitm_qual}
\end{figure*}
%------------------------------------------------------------------

\noindent \textbf{Impact of backbone model. }
We assessed the performance of scalp disease classification using different pretrained backbone models. 
Our evaluation metric was the F1 macro score, which we also used to compare our results with existing augmentation methods, including those utilizing DiffuseIT and AGG. 
The experiments involved two CNN-based models (ResNet~\cite{he2016:deep} and ResNeXt~\cite{xie2017:aggregated}) and two Transformer-based models (ViT~\cite{dosovitskiy2021:image} and EfficientFormer~\cite{li2022:efficientformer}), all maintaining consistent hyperparameters as described in Section~\ref{suppl:hyperparaemter_classification}.

As presented in Table~\ref{tab:sppl_quan_cls_backbone}, our approach achieved superior classification performance across all backbone models. 
Moreover, models augmented using generative methods consistently outperformed those using baseline augmentation, validating the effectiveness of our proposed data augmentation strategy.

\section{Qualitative Results of DIffuseIT-M}
\label{sec:suppl_diffitm}
Figure~\ref{fig:sppl_diffitm_qual} presents qualitative results of our model's ability to translate images across multiple labels while maintaining hair information. 
The results indicate that our model successfully translates various disease features from target images, effectively preserving hair representation.

\end{document}